\documentclass[10pt,twocolumn,letterpaper]{article}

\usepackage{cvpr}              

\usepackage[dvipsnames]{xcolor}

\definecolor{cvprblue}{rgb}{0.21,0.49,0.74}
\usepackage[pagebackref,breaklinks,colorlinks,citecolor=cvprblue]{hyperref}
\usepackage{float}
\usepackage{multirow}

\def\paperID{****} 
\def\confName{CVPR}
\def\confYear{2024}

\title{A General Framework for Jersey Number Recognition in Sports Video}

\author{Maria Koshkina\\
York University\\
Toronto, Canada\\
{\tt\small koshkina@yorku.ca}
\and
James H. Elder\\
York University\\
Toronto, Canada\\
{\tt\small jelder@yorku.ca}
}

\begin{document}
\maketitle

\begin{abstract}
Jersey number recognition is an important task in sports video analysis, partly due to its importance for long-term player tracking. It can be viewed as a variant of scene text recognition. However, there is a lack of published attempts to apply scene text recognition models on jersey number data. Here we introduce a novel public jersey number recognition dataset for hockey and study how scene text recognition methods can be adapted to this problem.  We address issues of occlusions and assess the degree to which training on one sport (hockey) can be generalized to another (soccer).  For the latter, we also consider how jersey number recognition at the single-image level can be aggregated across frames to yield tracklet-level jersey number labels.  We demonstrate high performance on image- and tracklet-level tasks, achieving 91.4\% accuracy for hockey images and 87.4\% for soccer tracklets. Code, models, and data are available at \href{https://github.com/mkoshkina/jersey-number-pipeline}{https://github.com/mkoshkina/jersey-number-pipeline}.
\end{abstract}

\section{Introduction}
\label{sec:intro}
Jersey number recognition is an important task in sports video understanding and automated game analysis.  One of the reasons it is so important is that sports video understanding depends fundamentally upon long-term tracking (over many minutes, i.e., many thousands of frames) of individual players.  Since players on the same team are dressed to look almost identical, the jersey number is a very precious feature that can serve to disambiguate tracks, especially across frames in which players become tightly clustered, as is common in many team sports.

Jersey number recognition can be a very challenging task, as the jersey number is typically only clearly visible on a minority of frames, and motion blur, body pose variations, projective distortions, occlusions, and folds in the jersey material causing complex distortions all conspire to make reliable recognition difficult.   Previous methods have approached the problem as a ground-up classification task, in which a network is trained from scratch.  A problem with this approach is that training the network from scratch requires a large labelled training dataset; Thus far, these have been proprietary and not released publicly.   Here we study whether the problem can be made more accessible by making use of Scene Text Recognition (STR) systems, pre-trained on more general large-scale synthetic and text-in-the-wild datasets.  We assess performance when using these systems out of the box and when first fine-tuning on a modest jersey number dataset.  We also assess how well such a system can generalize across sports, and very different camera geometries, with or without additional fine-tuning, and how best to aggregate image-level recognition to label tracklets comprised of many frames.  

A foundation of this research is a novel hockey jersey number dataset. It consists of hockey player images collected from university-level hockey games recorded from a stationary camera as well as hockey player images from the McGill NHL public tracking dataset\cite{MHPTD, zhao20}. The dataset has been manually annotated with the correct jersey number if it is legible by human eyes and a flag to indicate it is illegible otherwise.  

In summary, our main contributions are:
\begin{itemize}
\item A novel image-level dataset for hockey jersey number recognition.
\item A high-performance pipeline for detection, localization and frame-level recognition of jersey numbers. 
\item An analysis of how well this pipeline can generalize across sports, camera geometries and frame- vs tracklet-level jersey number recognition, with and without fine-tuning.
\end{itemize}

\section{Related Work}
\label{sec:related}

\subsection{Jersey Number Recognition}
The problem of jersey number recognition has been posed as image-level recognition \cite{liu2022jede, vats2021multi, bhargavi2022knock, liu2019pose, li2018jersey} as well as tracklet-level recognition \cite{vats2023player, vats2022ice, chan2021player, balaji2023jersey}.  Some methods detect and localize the jersey number region and then classify the numbers \cite{liu2022jede,liu2019pose,li2018jersey}, while others assume that the image region containing the jersey number has already been cropped \cite{vats2021multi, bhargavi2022knock, gerke2015soccer}.  

Progress on this problem has been slowed by the lack of large public datasets that can be used to compare methods.  This is now starting to be addressed with the 2023 release of the SoccerNet Jersey Number dataset \cite{Cioppa2022Scaling,SoccerNetJSChallenge}.

\subsubsection{Image-level Jersey Number Recognition}
Gerke et al. \cite{gerke2015soccer} and Li et al. \cite{li2018jersey} were among the first  to apply CNN-based classification approaches to image-level jersey number recognition, and CNNs have been the dominant approach since this time. 
Liu et al. \cite{liu2019pose,liu2022jede} demonstrated the utility of body pose detection to improve classification with Faster R-CNN \cite{ren2015faster} and Mask R-CNN \cite{he2017mask} architectures, respectively.   

Vats et al. \cite{vats2021multi} demonstrated that multi-task training of a network on both holistic and digit-wise number classification results in better performance than a network trained on either task alone. They made use of a large, labelled dataset but unfortunately it has not been made public.  Also, despite its size, the training dataset does not include all possible jersey numbers and the system cannot generalize to other numbers.  Bhargavi et al. \cite{bhargavi2022knock} employed a similar approach but pre-trained using synthetic data and then fine-tuned on a small, labelled dataset of real images. 

Nady et al. \cite{nady2021player} and Chen et al. \cite{chen2023tracking} explored using scene text detection and scene text recognition for jersey number recognition using CRAFT \cite{baek2019character}.  Although they show promising results, applying scene text detection involves fine-tuning text detector on jersey number bounding boxes. Thus, requiring additional annotation effort.

\subsubsection{Tracklet-level Jersey Number Recognition}
The visibility of a player's jersey number in video frames is often compromised due to motion blur and the player's position relative to the camera. In many instances, a player can be obscured by others, leading to multiple jersey numbers appearing in the same image. Identifying and pinpointing the jersey number of interest within a player's sequence of frames is a key step for recognizing jersey numbers at the tracklet level.  Vats et al.  \cite{vats2021multi} approach this by first classifying frames in each player's tracklet as legible or illegible. They then use only legible images to classify the number. On the other hand, Balaji et al. \cite{balaji2023jersey} propose a keyframe identification module that detects jersey numbers and filters out outliers (number detections that don't belong to the player in question or are too blurry for the recognition task). They use a jersey number detector and histogram-based features to detect and localize jersey numbers.  However, the specifics regarding any additional data or annotations used to fine-tune their detector remain unclear.  In our work, we take a similar approach to identify relevant images first.  Instead of relying on hand-crafted histogram-based features, our method utilizes features derived from a person re-identification network to filter out distractions, such as other players blocking the main subject. Similarly to \cite{vats2021multi}, we then apply a classifier to determine if the image contains legible numbers. Our approach is simple and shows superior performance on a challenging SoccerNet dataset.

For jersey number recognition at the tracklet level, there is an opportunity to integrate information across frames for better reliability.
Chan et al. \cite{chan2021player} and Balaji et al. \cite{balaji2023jersey} employed an LSTM while Vats et al. \cite{vats2023player} used a temporal convolutional network to aggregate information over time. Vats et al. \cite{vats2022ice} have also explored the use of transformers for tracklet-level jersey number recognition within the multi-task approach introduced in \cite{vats2021multi}.   They also make use of prior knowledge about the roster of players on the ice.

Most prior approaches treat jersey number recognition as a specialized classification problem requiring the design and training of a dedicated classification network.  In contrast, we propose to explore a system based upon a more generally trained scene text recognition (STR) model, which will allow our approach to take advantage of progressive improvements in STR technology, adapt to different scenarios with little or no fine-tuning, and handle all possible jersey numbers, instead of being restricted to numbers that happen to be in the training dataset.  In contrast to previously proposed jersey number STR approaches \cite{nady2021player, chen2023tracking} our method does not require jersey number bounding box annotations.  As in \cite{liu2019pose,liu2022jede}, we take advantage of body pose detection to localize the jersey number.  As in~\cite{vats2021multi} we use a weak-labelling strategy to generalize from image-level to tracklet-level annotation.  But in contrast to prior tracklet-level approaches~\cite{chan2021player,vats2021multi,vats2023player} we explore much simpler methods for integrating information across frames, demonstrating competitive results.

\subsection{Scene Text Recognition}
Scene Text Recognition (STR) is the task of recognizing text that occurs in the built environment (e.g., addresses, retail signs. traffic signs, license plates etc.).   Several large datasets containing both synthetic and real data have been made available to train STR models. The current state-of-the-art model PARSeq \cite{bautista2022parseq} uses an encoder and decoder architecture in addition to a learned language model. It shows high performance on several challenging real-world datasets that include character occlusions, diverse orientations and varied illumination. Due to the lack of image-level jersey number datasets, STR has not previously been trained or evaluated on the jersey number recognition task.  Here we explore how PARSeq can be integrated into a pipeline for jersey number recognition with and without fine-tuning.

\section{Method}

\label{sec:method}
\begin{figure*}[h]
\begin{center}
\includegraphics[width=0.9\linewidth]{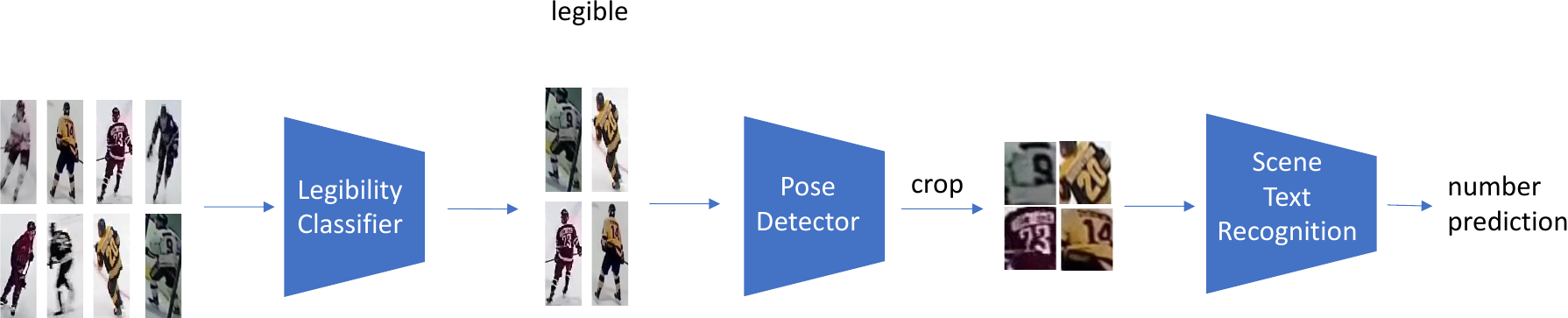}
\end{center}
\caption{Pipeline of image-level jersey number detection and recognition.}
\label{fig:pipeline}
\end{figure*}

\subsection{Overview}
To solve the jersey number recognition problem at the image level we introduce a simple yet very effective pipeline that detects, localizes and recognizes a jersey number of a player.  We then extend this pipeline to tracklet-level jersey number recognition by addressing challenges specific to that task: filtering out distractors and combining image predictions into a single tracklet-level prediction. We describe all these components in detail in subsequent sections.
\subsubsection{Image-level Task}
Figure~\ref{fig:pipeline} shows an overview of our image-level jersey number recognition pipeline.
In typical sports video, a jersey number is visible in only a minority of images.  Thus, the first step in jersey number recognition is to identify in which frames the number is visible and legible.  To perform this first task, we employ a binary CNN classifier based on an ImageNet\cite{ILSVRC15} pre-trained ResNet34\cite{he2016deep} model, fine-tuned on our new hockey dataset in which each player crop has been labelled as legible or illegible.  To estimate a bounding box around the jersey number we employ a body pose detector and use the estimated pose keypoints to crop out the player's torso region. To classify a jersey number within this bounding box we employ the state-of-the-art STR system PARSeq~\cite{bautista2022parseq} fine-tuned on a small number of hockey jersey number images.  
\subsubsection{Tracklet-level Task}
\begin{figure*}[h]
\begin{center}
\includegraphics[width=0.9\linewidth]{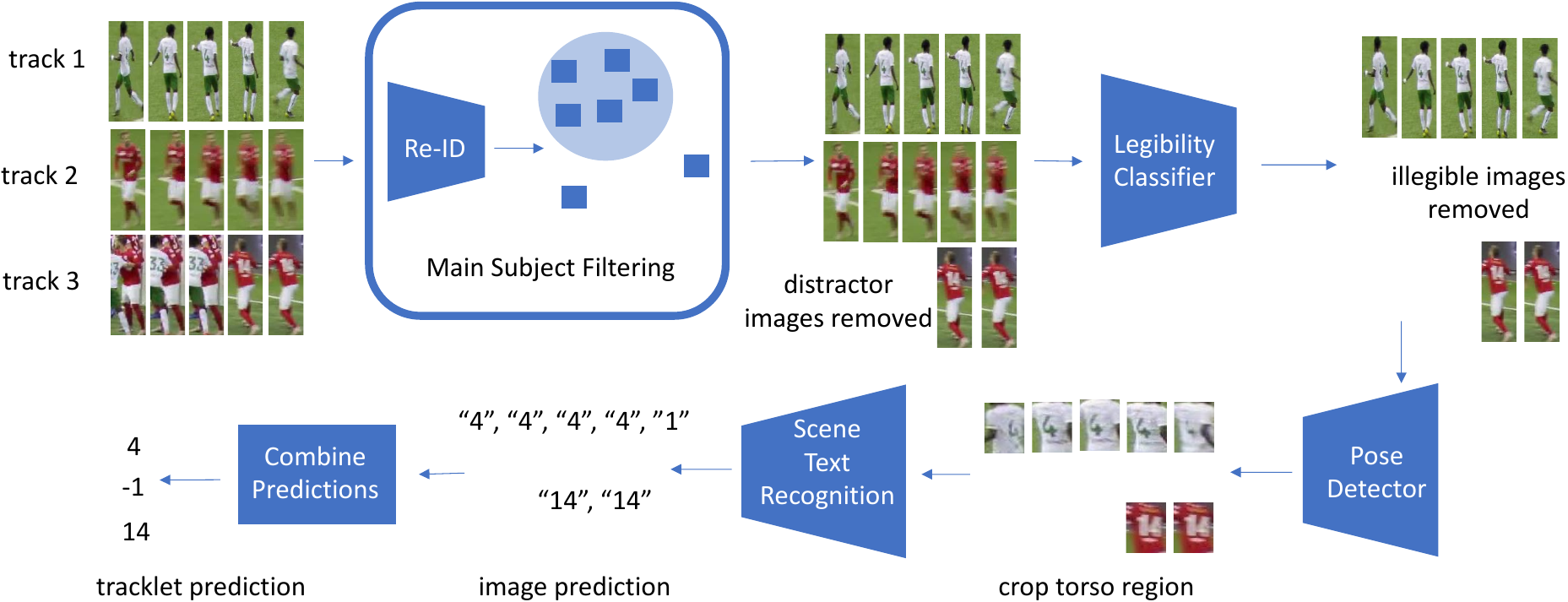}
\end{center}
\caption{Pipeline of tracklet-level jersey number detection and recognition.}
\label{fig:tracklet-pipeline}
\end{figure*}

To extend the above pipeline to the tracklet level  (Fig. \ref{fig:tracklet-pipeline}), we first use main subject filtering methods to identify frames that contain unoccluded players of interest. As in the image pipeline, we then employ our legibility classifier, followed by pose estimation to detect and localize jersey numbers. Finally, we use STR to recognize jersey numbers on each legible and unoccluded frame before aggregating image-level results over the entire tracklet.   

\subsection{Datasets}
To explore generalization across different sports, camera geometries, and image- vs tracklet-level classification, we employ two datasets:  Our own novel image-level hockey dataset and the recently-released tracklet-level SoccerNet soccer dataset \cite{Cioppa2022Scaling}. For both datasets, reliable jersey number recognition is challenging due to diversity in illumination, occlusions, motion blur, pose variations and material deformations. 

\begin{figure}[H]
\begin{center}
\includegraphics[width=0.9\linewidth]{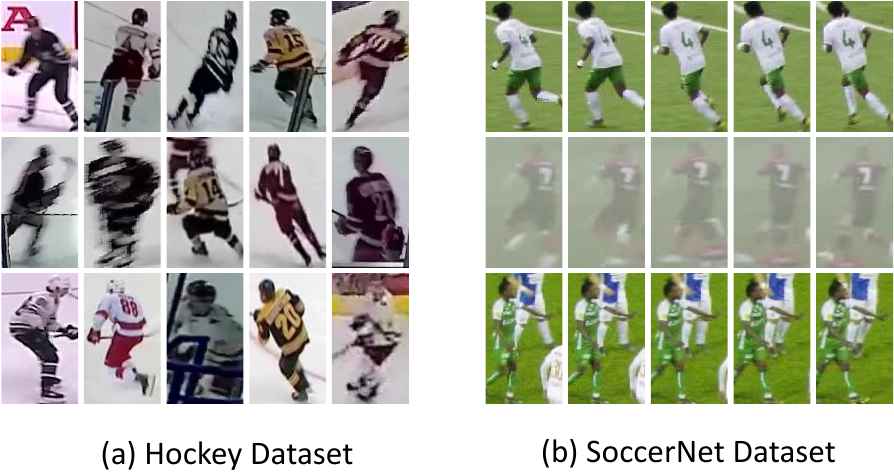}
\caption{Sample images from Hockey and SoccerNet datasets.}
\label{fig:dataset_samples}
\end{center}
\end{figure}

\subsubsection{Hockey}
To address the lack of publicly available image-level jersey number datasets, we introduce a new hockey jersey number dataset. We draw images from two sources:
\begin{itemize}
    \item University Hockey - player images from 9 different games recorded with a stationary camera.
    \item  McGill Hockey Player Tracking Dataset \cite{MHPTD, zhao20} - player images from 8 different NHL games from broadcast videos.
\end{itemize}

Note, that the camera geometries are very different.  While the University dataset is recorded with a fixed wide-field camera covering the whole rink, the McGill dataset is broadcast video, in which the camera zoom varies but is typically much more zoomed-in than for the university dataset.  For both datasets, there is a lot of motion blur and partial occlusion.  The university hockey images are especially challenging:  A single camera device captures the whole rink and there is no pan and zoom, so player images are typically of lower resolution and jersey numbers are harder to decipher.  To make a more diverse hockey dataset we combine images from both the university and NHL into a single labelled image-level jersey number dataset.  The data is available from \href{https://github.com/mkoshkina/jersey-number-pipeline}{https://github.com/mkoshkina/jersey-number-pipeline}.

The hockey image-level dataset consists of cropped player images and has two types of annotation: legibility and jersey number. Player images were labelled as legible if the annotator could be certain of the jersey number.  For jersey number recognition we used only a subset of these legible images to avoid excessive duplication of the same number. These images are labelled with a jersey number. We partitioned the data into training (10 games), validation (1 game) and test (6 games) - Table~\ref{table:hockey_dataset} details how this breaks down in terms of number of labelled images. Sample images from the dataset are shown in Figure~\ref{fig:dataset_samples}.

\begin{table}
  \centering
  \begin{tabular}{{|l|c|c|c|}}
    \hline
    Part & \multicolumn{2}{c|}{Legibility} & Jersey Number\\
    \hline
      & legible & total & \\
      \hline
      \hline
     Train & 4,706 & 94,036  & 3,531 \\
     Validation & 923 & 14,138 & 233 \\
     Test & 2,158 & 24,809  & 486 \\
     \hline
     Total & 7,787 & 132,983 & 4,250 \\
    \hline
  \end{tabular}
  \caption{Hockey Dataset: number of labelled images by partition and annotation type.}
  \label{table:hockey_dataset}
\end{table}

\begin{figure*}
\begin{center}
\centering
\begin{subfigure}{.5\textwidth}
  \centering
  \includegraphics[width=.75\linewidth]{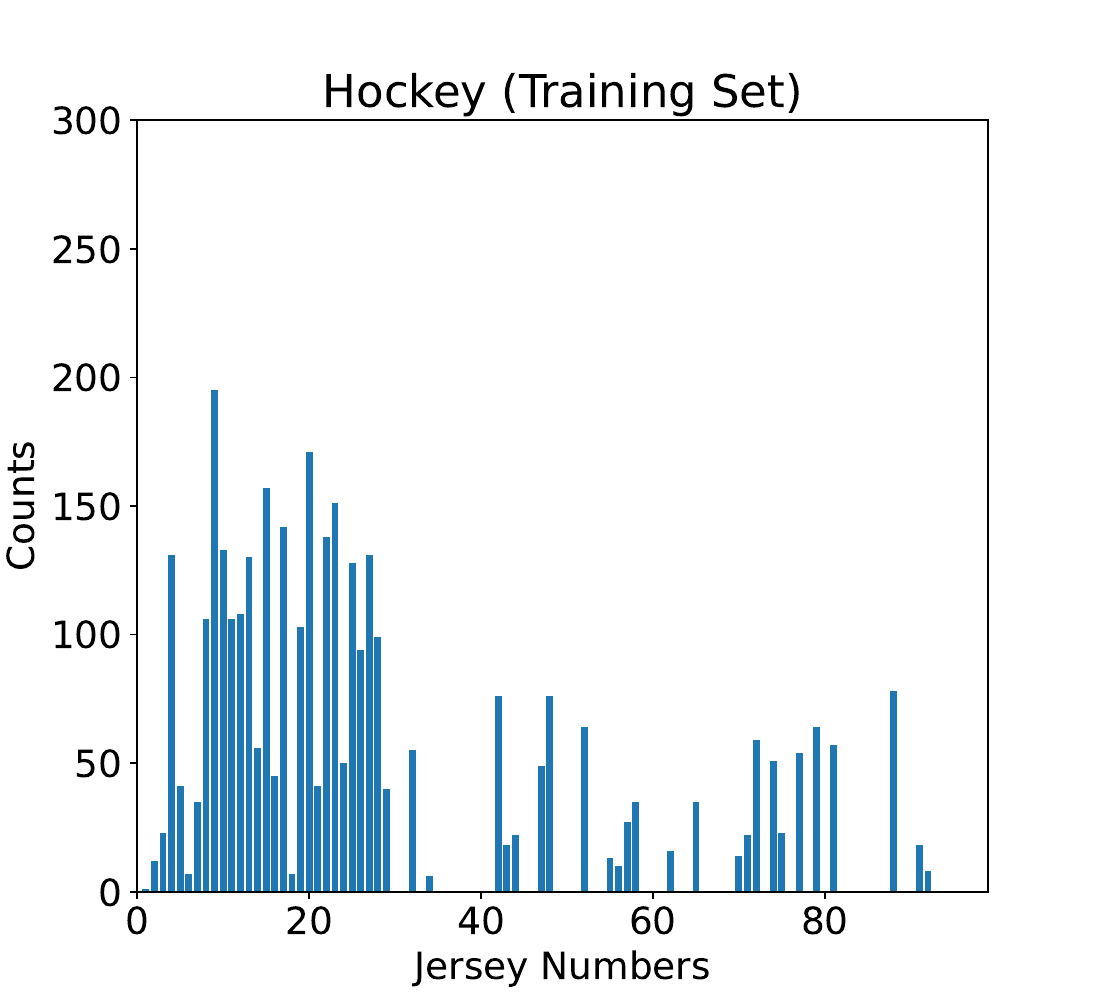}
  \label{fig:hockey_train}
\end{subfigure}%
\begin{subfigure}{.5\textwidth}
  \centering
  \includegraphics[width=0.75\linewidth]{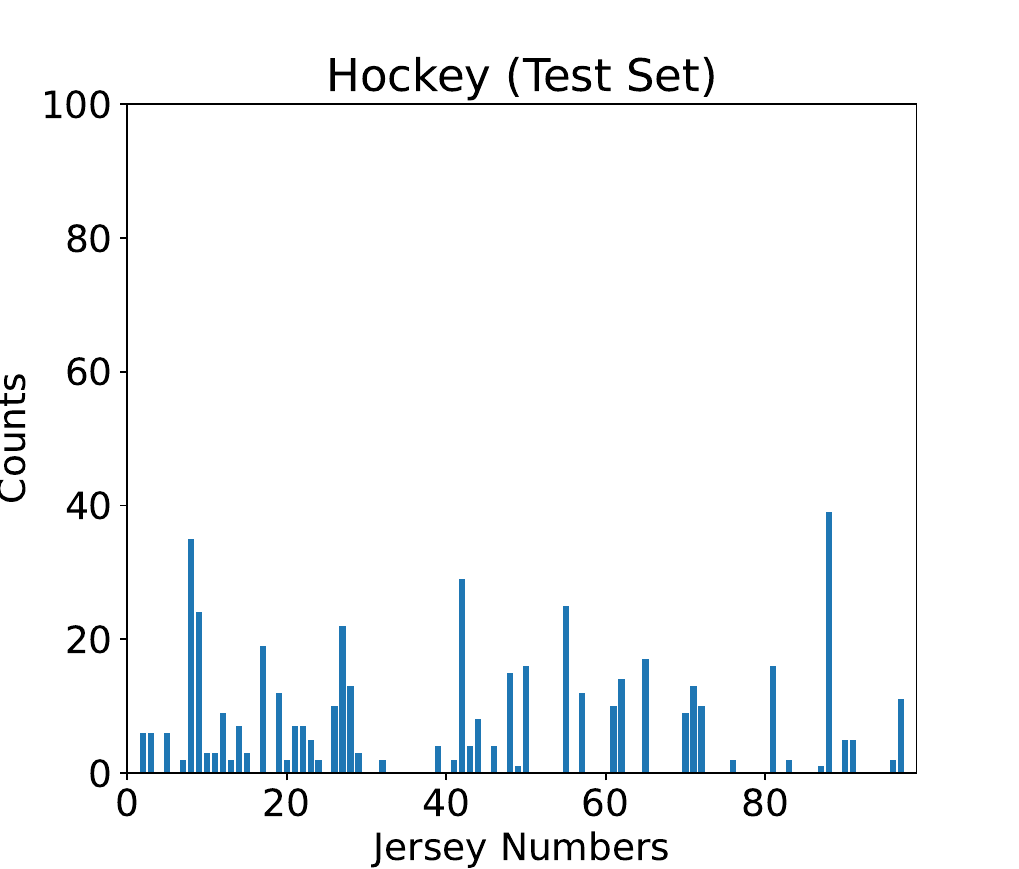}
  \label{fig:hockey_test}
\end{subfigure}
\end{center}
\caption{Hockey Dataset jersey number distribution.}
\label{fig:hockey_ds_stats}
\end{figure*}

Figure~\ref{fig:hockey_ds_stats} shows the distribution of jersey numbers for training and test. There are 54 unique jersey numbers in the training set and 25 in the test set.  Two numbers in the test set do not appear in the training set.

\subsubsection{Soccer}
In early 2023, SoccerNet~\cite{Cioppa2022Scaling} released the Jersey Number Recognition dataset and challenge \cite{SoccerNetJSChallenge}, making it the first large public jersey number recognition dataset.  This dataset consists of a collection of player tracklets and contains tracklet-level annotation.  The dataset is partitioned into training, test and challenge, with a total of 4,064 tracklets. The average length of tracklets is 482 frames. Table~\ref{table:soccer_dataset} contains dataset statistics and Figure~\ref{fig:dataset_samples} shows sample images. During our experiments we discovered a flaw in the annotations:  it includes tracklets for a soccer ball with the label of jersey number "1". We added a component to our pipeline to identify soccer ball detections based on the average dimensions of the soccer ball images in the training set.  

\begin{figure*}
\begin{center}
\centering
\begin{subfigure}{.5\textwidth}
  \centering
  \includegraphics[width=.75\linewidth]{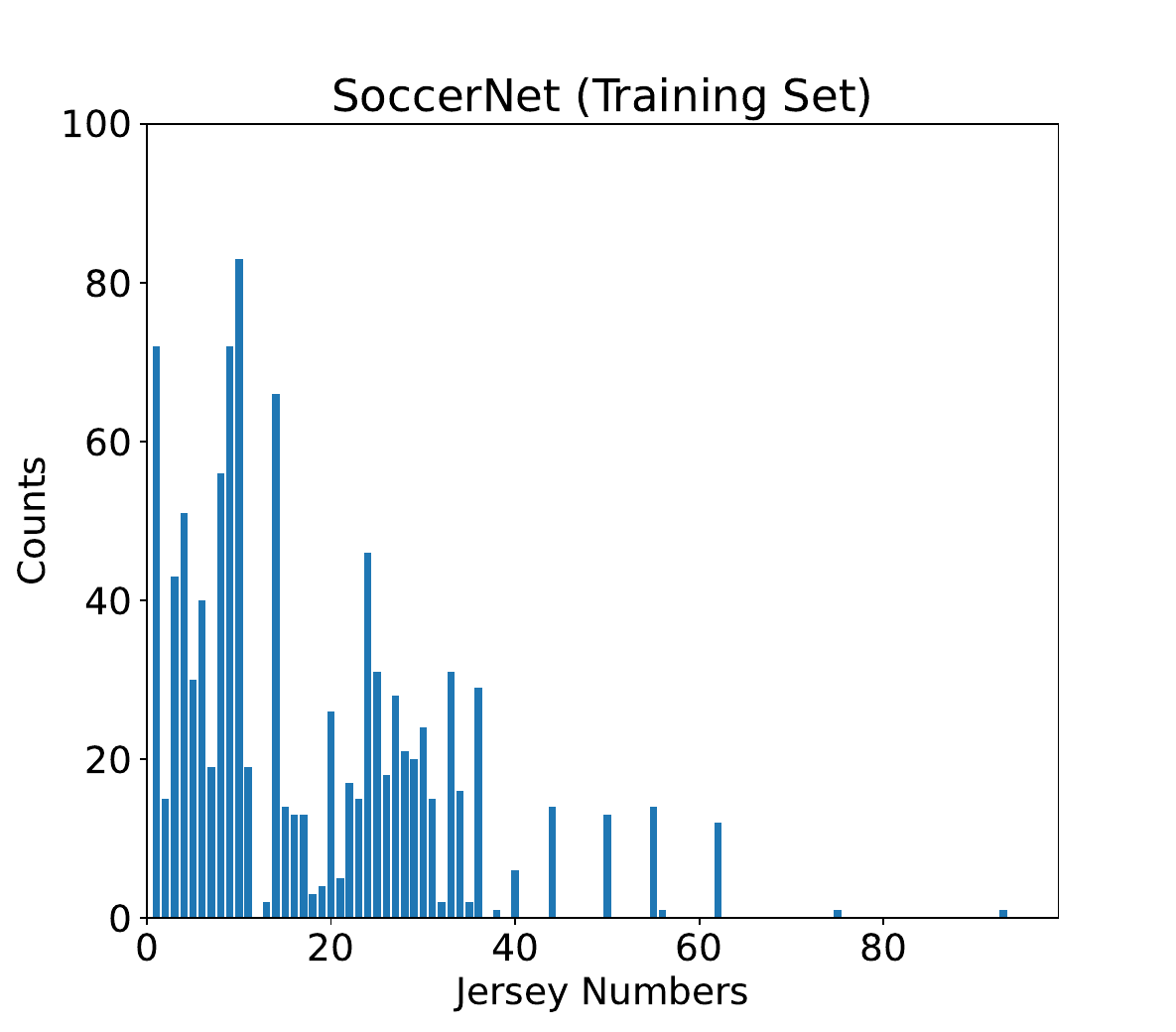}
  \label{fig:soccer_train}
\end{subfigure}%
\begin{subfigure}{.5\textwidth}
  \centering
  \includegraphics[width=0.75\linewidth]{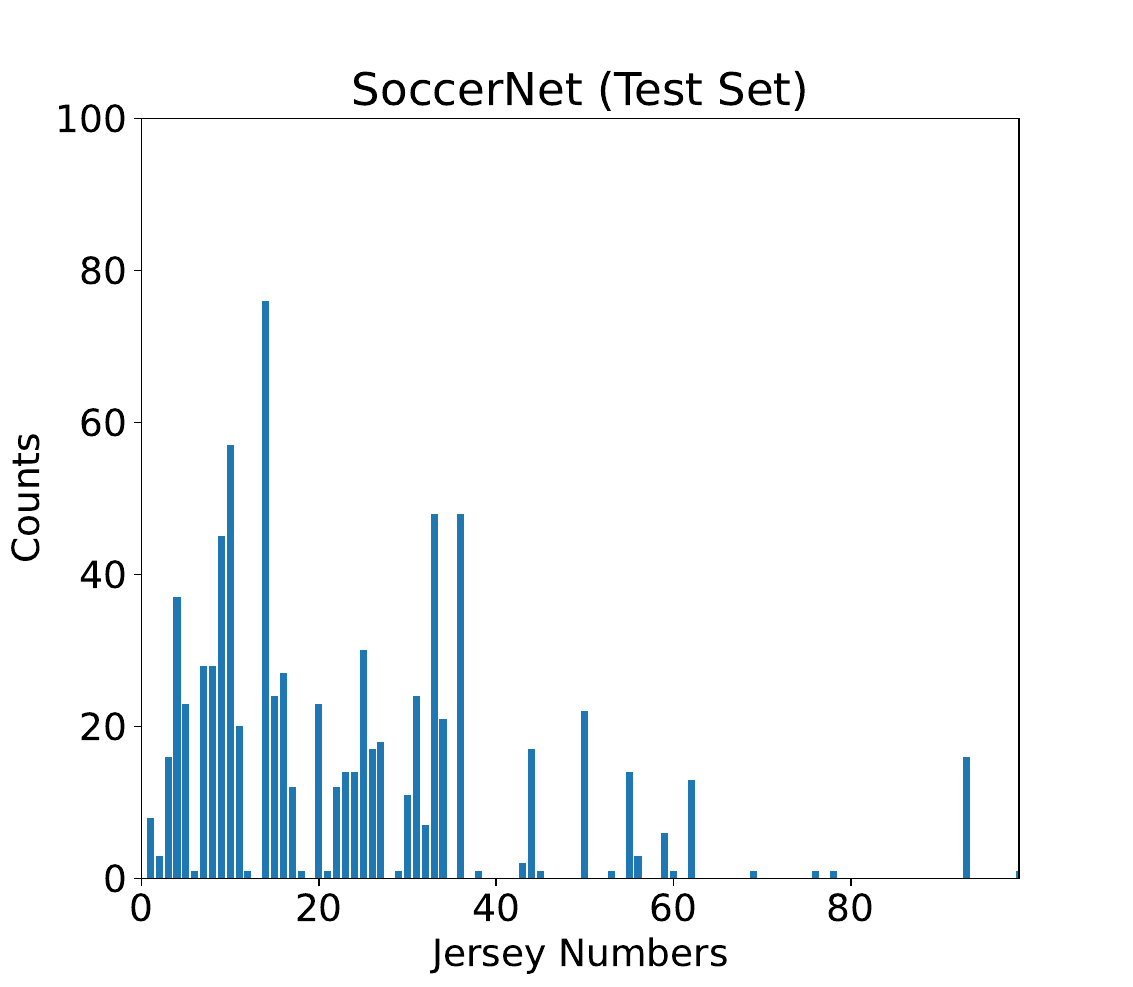}
  \label{fig:soccer_test}
\end{subfigure}
\end{center}
\caption{SoccerNet Dataset jersey number distribution.}
\label{fig:soccer_ds_stats}
\end{figure*}

There are 55 unique jersey numbers in the training and test partitions and the test set contains 10 numbers that do not appear in the training partition. Figure ~\ref{fig:soccer_ds_stats} shows their distribution.

\begin{table}
\begin{center}
\begin{tabular}{|l|c|c|c|c|}
\hline
 & Train & Test & Challenge & Total  \\
\hline\hline
Tracklets & 1,427 & 1,211 & 1,426 & 4,064 \\
Images & 733K & 564.5K &  748.6K & 2,046K \\
\hline
\end{tabular}
\end{center}
\caption{SoccerNet Jersey Number Dataset.}
\label{table:soccer_dataset}
\end{table}

\subsection{Image-level Task}
\subsubsection{Detection and Localization}
A jersey number is typically only visible and legible on a fraction of the frames.  To filter out images with illegible numbers we train a binary classifier to identify player images as either legible (has a visible and decipherable jersey number) or illegible.  Since jersey numbers are located on the torso of the player, we  utilize a pose estimator to extract the torso region.  This approach is simpler than training a dedicated jersey number detector because it does not require time-consuming jersey number bounding box annotation. Instead, it relies on a simple legible/illegible binary label and an off-the-shelf pose estimation network. 

For our legibility classifier we employ a ResNet34~\cite{he2016deep} model pre-trained on ImageNet~\cite{ILSVRC15} and fine-tune it on our binary hockey legibility dataset. Our hockey legibility dataset is highly imbalanced with only 5\% of images labelled as legible. Although this reflects the true distribution, our experiments showed that using a balanced training dataset  improved classifier performance. Therefore, we train with a balanced subset consisting of all legible images and an equal number of randomly selected  illegible images.  Test results are reported on the original imbalanced test data.

We train our binary legibility classifier for 20 epochs with a starting learning rate of 0.001 and momentum of 0.9. To improve the generalizability of our classifier we use Sharpness-Aware Minimization (SAM) \cite{foret2021sharpnessaware} with SGD. We provide a careful ablation study of legibility model choice, as well as model generalizability analysis in Section~\ref{sec:results}.  
 
\begin{figure}[t]
  \centering
\includegraphics[width=0.8\linewidth]{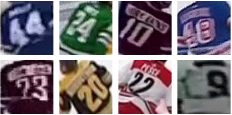}
\caption{Sample jersey number crops automatically extracted from player images.}
\label{fig:crops}
\end{figure}

We localize jersey number on the player image by extracting body pose keypoints using off-the-shelf body pose detector ViTPose\cite{xu2022vitpose} trained on MS COCO\cite{lin2014microsoft}. We then crop a rectangle defined by shoulder and hip joints padded by 5 pixels on the left, right, and bottom.  A sample of the resulting crops from our hockey dataset is shown in Figure~\ref{fig:crops}.

\subsubsection{Recognition}
Jersey number recognition is a specific case of Scene Text Recognition (STR).  Recent STR models show very good performance recognizing text in the wild. We fine-tune leading STR model PARSeq \cite{bautista2022parseq} to recognize jersey numbers. PARSeq is trained on a collection of synthetic and real-world datasets including SynthText\cite{gupta2016synthetic}, COCO-Text\cite{veit2016coco}, and TextOCR\cite{singh2021textocr} (refer to \cite{bautista2022parseq} for a full list of training datasets.) There are several advantages to relying on the existing STR model for this task. It has been pre-trained on a vast number of images containing alphanumeric strings.  It performs reasonably well on jersey number recognition tasks without any fine-tuning. Performance is further improved by fine-tuning on relatively small amount of jersey number data (see Table~\ref{table:hockey_dataset}).  Due to its token-processing nature, the model can predict jersey numbers that were not present in the training set, making it better suited to real-world applications. We fine-tune the model on legible jersey number crops from the hockey dataset for 25 epochs, limiting label length to 2 and using default PARSeq training settings.

Our proposed pipeline is simple, yet it outperforms previous methods. In the future, it can also benefit from advances in STR methods.

\subsection{Tracklet-level Task}
We adapt our image-level pipeline to the tracklet level  by introducing two additional  steps: main subject filtering and jersey number prediction consolidation. 

\subsubsection{Main Subject Filtering}
The SoccerNet dataset contains tracklets where the main subject is often occluded by other players.  When the jersey number of the occluding player is visible it can affect both legibility and number predictions for the tracklet.  This renders images where the main subject is occluded or multiple players are visible problematic.  We study whether filtering out frames in which the main subject appears to be occluded can improve tracklet-level classification. To this end, we employ the Centroid-ReID \cite{wieczorek2021unreasonable} network trained on the Market1501 dataset \cite{zheng2015scalable} to extract a visual feature vector for each image in a tracklet. We  fit an isotropic Gaussian to these vectors, and then exclude as outliers any images for which the feature vector lies more than N standard deviations  from the mean.  This process is repeated K times.  In our experiments, this method leads to better overall results.  Parameters for N and K were determined by grid search and cross-validation on a held out 30\% subset of the training set tracklets.  We found the optimal parameters to be $K=3$, $N=3.5$.   Note, that the method is unsupervised; there are no labels for the main subject in the tracklet. We evaluate its performance based on its impact on the tracklet-level recognition task.  Experiments show that this method leads to a boost in performance on the SoccerNet dataset (Section \ref{sec:results}).  

\subsubsection{Detection and Localization}
Extending our legibility classifier to the tracklet-level SoccerNet jersey recognition task is complicated by the lack of frame-by-frame labels.  To overcome this barrier, we derive weak pseudo-labels from the tracklet-level annotations.   We derive a set of positive (legible) pseudo-label instances by running our hockey-trained legibility classifier on the images within legible tracklets (tracklets with jersey number labels) and extracting instances deemed legible.  Negatives are drawn from random images from illegible tracklets.  We train the legibility classifier network using these pseudo-labels. 
At inference, a tracklet is deemed legible if it contains one or more images classified as legible.

\subsubsection{Recognition}
\label{subsection:recognition}
To fine-tune STR for the tracklet-level SoccerNet dataset we construct a weakly-labelled text recognition dataset based on tracklet-level data. In particular, we run our legibility classifier on all images in legible tracklets (tracklets with a jersey number label) and use these as pseudo-ground truth for fine-tuning.  

At inference, we run the fine-tuned STR model on all images deemed legible in the tracklet. The result is a series of predicted jersey number labels: one for each legible image in the tracklet.

\subsubsection{Prediction Consolidation}
\begin{figure}[th]
\begin{center}
\includegraphics[width=0.4\linewidth]{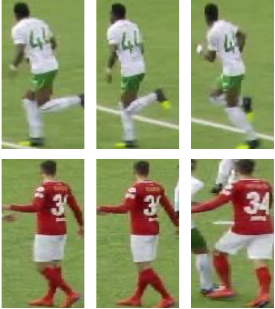}
\end{center}
\caption{Example of images where only one digit out of the two is visible.  First row: true label 44, predicted 4. Second row: true label 34, predicted 3.}
\label{fig:digit-bias}
\end{figure}

\begin{table}
\begin{center}
\begin{tabular}{l|l|c|c|}
\multicolumn{2}{c}{}&\multicolumn{2}{c}{Ground Truth}\\
\cline{3-4}
\multicolumn{2}{c|}{}&2 digit&1 digit\\
\cline{2-4}
\multirow{2}{*}{Prediction}& 2 digit & 40\% & 7\% \\
\cline{2-4}
& 1 digit & \textbf{48\%} & 5\% \\
\cline{2-4}
\end{tabular}
\end{center}
\caption{Confusion matrix of STR predictions with regard to one- or two-digit jersey numbers.}
\label{table:str_errors}
\end{table}
We investigated two distinct approaches to consolidating individual image predictions into a tracklet-level prediction:  One heuristic and one probabilistic.  An important consideration when approaching this problem is the potential confusion between one- and two-digit jersey numbers.  Two-digit jersey numbers are roughly twice as frequent as one-digit numbers in the SoccerNet dataset. Due to occlusions and variations in player pose, only one digit may be visible even when the jersey number consists of two digits  (Figure~\ref{fig:digit-bias}).  As a result, STR confusion regarding the number of digits in the jersey number is overwhelmingly due to mistaking a 2-digit number for a 1-digit number (Table~\ref{table:str_errors}).\\
\textbf{Heuristic consolidation:} In our heuristic approach, we compute the tracklet-level prediction using a confidence-weighted majority vote of legible images. If the sum of confidences over frames is below a threshold, the tracklet is marked illegible.  When some of the images in the tracklet are predicted to have two digits and others to have a single digit, we down-weight votes for one-digit numbers. Both of these measures provide a small boost to overall performance.\\
\textbf{Probabilistic consolidation:}  In our probabilistic approach, we consider the one or two digits of the jersey number separately.
For each position in the string, the STR system outputs a vector of length $K+1$ where $K$ is the number of characters in the language and there is an additional special character that indicates the end of the string. This vector represents predicted probabilities for each of the characters in the language being in this specific position in the string. For our jersey number prediction maximum string length is 2 and $K=10$. Therefore, STR outputs 2 vectors of length 11.  We assume a uniform prior over digits 0-9. Only 39\% of numbers in our dataset are single-digit. We incorporate this bias into prior probability of seeing 'end-of-string' character in the 2nd position.  

To address the (typical) overconfidence of the STR network, we apply a standard temperature scaling algorithm \cite{guo2017calibration} to recalibrate these confidences to better reflect true posterior probabilities.  We denote the likelihoods as $p\left(I_n|d_{j}=k\right)$ of observing image $I_n$ given the existence of character $k$ in position $j$ with prior of $p\left(d_{j}=k\right)$.  Assuming conditional independence over time, we compute the sum of log-likelihoods for each digit position over all legible images in the tracklet:
\begin{eqnarray}
\begin{split}
p\left(d_{j}=k|\left\{I_n\right\}\right)=\sum_{n\in\mathcal{N}_l}(\log p\left(I_n|d_{j}=k\right)+\\ \log p\left(d_{j}=k\right))
\end{split}
\end{eqnarray}
where $\mathcal{N}_l$ is the set of legible images in the tracklet.  

The predicted value of digit $j$ is then
$\arg\max_k p\left(d_{j}=k|\left\{I_n\right\}\right)$.  
 
\section{Results and Analysis}
\label{sec:results}
\subsection{Image-Level Task}
Our ResNet34\cite{he2016deep} legibility classifier performs at 94.5\% accuracy with F1-score of 71.7\% on our hockey test set.  We also evaluated a fine-tuned visual transformer model~\cite{dosovitskiy2020image} (See Table~\ref{table:legible_results}) but found that, while it performs better when tested on the same dataset it was trained on, ResNet34\cite{he2016deep} shows better results in generalizing to the new domain. 

\begin{table}
\begin{center}
\begin{tabular}{|l|c|c|}
\hline
Method & Dataset Size & Accuracy  \\
\hline\hline
Li et al.  \cite{li2018jersey} & 12,746 & $86.7\%$ \\
Liu et al.  \cite{liu2019pose} & 3,567 & $90.4\%$ \\
Vats et al. \cite{vats2021multi}   & 54,251 & $89.6\%$\\
Bhargavi et al. \cite{bhargavi2022knock} & 3,000 & $89.3\%$\\
\hline
\textbf{Ours} & 4,250 & $\boldsymbol{91.4\%}$ \\
\hline
\end{tabular}
\end{center}
\caption{Previously reported results on image-level jersey number recognition task.}
\label{table:hockey_comparison}
\end{table}
\begin{table}
\begin{center}
\begin{tabular}{|l|c|}
\hline
Model & Accuracy  \\
\hline\hline
Holistic Classifier (ResNet34) & $48.1\%$ \\
Multi-Task Classifier (ResNet34) \cite{vats2021multi}  & $65.2\%$ \\
PARSeq (out-of-the-box) \cite{bautista2022parseq}  & $85.4\%$\\
\hline
\textbf {Ours:} PARSeq (fine-tuned on hockey) & $\boldsymbol{91.4\%}$\\
\hline
\end{tabular}
\end{center}
\caption{Performance of jersey number recognition models on our hockey dataset.}
\label{table:hockey_recognition_results}
\end{table}
\begin{table}
\begin{center}
\begin{tabular}{|l|c|c|}
\hline
 & Test: Hockey & Test: Soccer  \\
\hline\hline
Original & $85.40\%$ & $80.51\%$ \\
Fine-tune: Hockey & ${91.40\%}$ & $83.90\%$ \\
Fine-tune: Soccer & $65.84\%$ & $87.45\%$ \\
\hline
\end{tabular}
\end{center}
\caption{Generalizability of the PARSeq STR model on hockey and soccer datasets with and without fine-tuning.}
\label{table:str_generalizability}
\end{table}
\begin{table*}
\begin{center}
\begin{tabular}{|l|c|c|c|c|c|c|c|c|}
\hline
Model &  \multicolumn{2}{c|}{$H \rightarrow H$} & \multicolumn{2}{c|}{$H \rightarrow S$} & \multicolumn{2}{c|}{$S \rightarrow S$} & \multicolumn{2}{c|}{$S \rightarrow H$} \\
\hline
 & Acc & F1 & Acc & F1 & Acc & F1 & Acc & F1 \\
\hline
\hline
ResNet18 \cite{he2016deep} & $94.8\%$ & $71.4\%$  & $90.58\%$ & $93.0\%$  & $91.71\%$ & $94.15\%$  & $91.9\%$ & $65.3\%$ \\
ResNet34 \cite{he2016deep}  & $94.5\%$ & $71.7\%$   & $91.09\%$ & $93.7\%$  & $91.71\%$ & $94.17\%$  & $92.8\%$ & $63.2\%$\\
ViT \cite{dosovitskiy2020image}  & $94.8\%$ & $72.9\%$  & $86.9\%$ & $90.5\%$  & $90.75\%$ & $93.6\%$  & $92.6\%$ & $58.3\%$\\
\hline
\end{tabular}
\end{center}
\caption{Performance comparison of three deep architectures for our legibility classifier.  We examine how well different models generalize from $(Training Dataset) \rightarrow (Testing Dataset)$ and report both accuracy and F1 scores. Accuracy for Soccer is calculated at the tracklet level (a tracklet is deemed legible if it contains one or more legible images).}
\label{table:legible_results}
\end{table*}
\begin{table*}
\begin{center}
\begin{tabular}{|l|c|c|c|c|}
\hline
Consolidation & Full &  No Bias & No Bias, No Threshold & No Filtering \\
\hline\hline
Probabilistic & $85.22\%  (\downarrow 2.23\%)$ & $85.05\%  (\downarrow 2.40\%)$ & - & -\\
Heuristic & $\boldsymbol{87.45\%}$ & $86.79\%  (\downarrow 0.66\%)$ & $85.38\%(\downarrow 2.07\%)$ & $84.56\% (\downarrow 2.89\%)$\\
\hline
\end{tabular}
\end{center}
\caption{Ablation analysis of our soccer pipeline.  We consider heuristic or probabilistic consolidation methods, with or without biasing toward two-digit jersey numbers.  For the heuristic merthod we also evaluate the effect of placing a threshold on the sum of confidences.   Final column shows the performance of the heuristic method with no main subject filtering.  The results are on SoccerNet test set.}
\label{table:soccer_ablation}
\end{table*}
\begin{table}[t]
\begin{center}
\begin{tabular}{|l|c|c|}
\hline
Method & Test Acc & Challenge Acc \\
\hline\hline
Gerke et al \cite{gerke2015soccer} & 32.57\% & 35.79\% \\
Vats et al \cite{vats2021multi} & 46.73\% & 49.88\% \\
Li et al \cite{li2018jersey} & 47.85\% & 50.60\%\\
Vats et al \cite{vats2022ice} & 52.91\% & 58.45\%\\
Balaji et al \cite{balaji2023jersey} & 68.53\% & 73.77\%\\
\hline
\textbf{Ours} & \textbf{87.45\%} & \textbf{79.31\%}\\
\hline
\end{tabular}
\end{center}
\caption{Tracklet-level jersey number recognition performance on the SoccerNet Test and Challenge partitions. Results for other methods are cited from \cite{balaji2023jersey}.}
\label{table:soccer_results}
\end{table}
We evaluate jersey number recognition on image-level annotations for our hockey dataset considering only legible images and achieve an accuracy of 91.4\%. Table~\ref{table:hockey_comparison} shows a comparison with methods previously reported in the literature. As a baseline, we evaluated both a ResNet34-based classifier trained on our data to predict a label 1-99, as well as the multi-task system described in \cite{vats2021multi} that uses a holistic classifier and digit-wise classifier heads. As with the results reported in \cite{vats2021multi}, this multi-task training yields better results, but due to our small training set we see much lower performance than Vats et al. \cite{vats2021multi} reported.  Without any fine-tuning PARSeq \cite{bautista2022parseq} trained on multiple synthetic and real scene text datasets achieves an accuracy of 85.4\% on our hockey image-level dataset. The performance further improves with fine-tuning illustrating that the use of STR in the jersey number recognition pipeline is an appropriate choice (Table~\ref{table:hockey_recognition_results}).

\subsection{Tracklet-Level Task}
To evaluate recognition on the tracklet-level SoccerNet dataset we use the evaluation protocol followed in the SoccerNet Jersey Number Recognition Challenge.
We evaluate the accuracy of tracklet-level labelling in which each tracklet may be comprised of both legible and illegible frames. Using the full pipeline with a legibility classifier and fine-tuned PARSeq model we achieve an accuracy of 87.45\% on the SoccerNet test set and 79.31\% on the challenge set. Table~\ref{table:soccer_results} shows the results of our method compared to previously published on this dataset.  Perhaps, due to the complexity of the hockey dataset, the PARSeq model performs better on soccer when trained on hockey than vice versa (Table~\ref{table:str_generalizability}).

In Table~\ref{table:soccer_ablation} we present the results of several ablations.  In particular, we demonstrate the effect of main subject filtering as well as different options for prediction consolidation. Our best results are achieved using the heuristic consolidation approach.

\section{Conclusions \& Future Work}
We have introduced a robust pipeline designed for jersey number recognition at both image and tracklet levels. Our system outperforms previously reported results while requiring minimum fine-tuning. It generalizes exceptionally well to new jersey numbers as well as from one sport to another. 

Furthermore, in an effort to foster continued research and development in this domain, we have introduced a novel dataset for image-level recognition of hockey jersey numbers.

Looking ahead, we envision integrating jersey number recognition into player tracking. This integration has the potential to significantly enhance player tracking systems, providing richer and more comprehensive data for sports analytics and improving overall performance in player monitoring and analysis.

{
    \small
    \bibliographystyle{ieeenat_fullname}
    \bibliography{jersey_numbers}

\begin{thebibliography}{32}
\providecommand{\natexlab}[1]{#1}
\providecommand{\url}[1]{\texttt{#1}}
\expandafter\ifx\csname urlstyle\endcsname\relax
  \providecommand{\doi}[1]{doi: #1}\else
  \providecommand{\doi}{doi: \begingroup \urlstyle{rm}\Url}\fi

\bibitem[MHP()]{MHPTD}
{McGill Hockey Player Tracking Dataset (MHPTD)}.
\newblock \url{https://github.com/grant81/hockeyTrackingDataset}.

\bibitem[Soc()]{SoccerNetJSChallenge}
{SoccerNet} {Jersey Number Recogntion}.
\newblock \url{https://www.soccer-net.org/tasks/jersey-number-recognition}.

\bibitem[Baek et~al.(2019)Baek, Lee, Han, Yun, and Lee]{baek2019character}
Youngmin Baek, Bado Lee, Dongyoon Han, Sangdoo Yun, and Hwalsuk Lee.
\newblock Character region awareness for text detection.
\newblock In \emph{Proceedings of the IEEE/CVF Conference on Computer Vision
  and Pattern Recognition}, pages 9365--9374, 2019.

\bibitem[Balaji et~al.(2023)Balaji, Bright, Prakash, Chen, Clausi, and
  Zelek]{balaji2023jersey}
Bavesh Balaji, Jerrin Bright, Harish Prakash, Yuhao Chen, David~A Clausi, and
  John Zelek.
\newblock Jersey number recognition using keyframe identification from
  low-resolution broadcast videos.
\newblock In \emph{Proceedings of the 6th International Workshop on Multimedia
  Content Analysis in Sports}, pages 123--130, 2023.

\bibitem[Bautista and Atienza(2022)]{bautista2022parseq}
Darwin Bautista and Rowel Atienza.
\newblock Scene text recognition with permuted autoregressive sequence models.
\newblock In \emph{European Conference on Computer Vision}, pages 178--196,
  Cham, 2022. Springer Nature Switzerland.

\bibitem[Bhargavi et~al.(2022)Bhargavi, Coyotl, and Gholami]{bhargavi2022knock}
Divya Bhargavi, Erika~Pelaez Coyotl, and Sia Gholami.
\newblock Knock, knock. who's there?--identifying football player jersey
  numbers with synthetic data.
\newblock \emph{arXiv preprint arXiv:2203.00734}, 2022.

\bibitem[Chan et~al.(2021)Chan, Levine, and Javan]{chan2021player}
Alvin Chan, Martin~D Levine, and Mehrsan Javan.
\newblock Player identification in hockey broadcast videos.
\newblock \emph{Expert Systems with Applications}, 165:\penalty0 113891, 2021.

\bibitem[Chen and Poullis(2023)]{chen2023tracking}
Qiao Chen and Charalambos Poullis.
\newblock Tracking and identification of ice hockey players.
\newblock In \emph{International Conference on Computer Vision Systems}, pages
  3--16. Springer, 2023.

\bibitem[Cioppa et~al.(2022)Cioppa, Deli{\`e}ge, Giancola, Ghanem, and
  Droogenbroeck]{Cioppa2022Scaling}
Anthony Cioppa, Adrien Deli{\`e}ge, Silvio Giancola, Bernard Ghanem, and
  Marc~Van Droogenbroeck.
\newblock Scaling up {SoccerNet} with multi-view spatial localization and
  re-identification.
\newblock \emph{Scientific Data}, 9, 2022.

\bibitem[Dosovitskiy et~al.(2020)Dosovitskiy, Beyer, Kolesnikov, Weissenborn,
  Zhai, Unterthiner, Dehghani, Minderer, Heigold, Gelly,
  et~al.]{dosovitskiy2020image}
Alexey Dosovitskiy, Lucas Beyer, Alexander Kolesnikov, Dirk Weissenborn,
  Xiaohua Zhai, Thomas Unterthiner, Mostafa Dehghani, Matthias Minderer, Georg
  Heigold, Sylvain Gelly, et~al.
\newblock An image is worth 16x16 words: Transformers for image recognition at
  scale.
\newblock \emph{arXiv preprint arXiv:2010.11929}, 2020.

\bibitem[Foret et~al.(2021)Foret, Kleiner, Mobahi, and
  Neyshabur]{foret2021sharpnessaware}
Pierre Foret, Ariel Kleiner, Hossein Mobahi, and Behnam Neyshabur.
\newblock Sharpness-aware minimization for efficiently improving
  generalization.
\newblock In \emph{International Conference on Learning Representations}, 2021.

\bibitem[Gerke et~al.(2015)Gerke, Muller, and Schafer]{gerke2015soccer}
Sebastian Gerke, Karsten Muller, and Ralf Schafer.
\newblock Soccer jersey number recognition using convolutional neural networks.
\newblock In \emph{Proceedings of the IEEE International Conference on Computer
  Vision Workshops}, pages 17--24, 2015.

\bibitem[Guo et~al.(2017)Guo, Pleiss, Sun, and Weinberger]{guo2017calibration}
Chuan Guo, Geoff Pleiss, Yu Sun, and Kilian~Q Weinberger.
\newblock On calibration of modern neural networks.
\newblock In \emph{International Conference on Machine Learning}, pages
  1321--1330. PMLR, 2017.

\bibitem[Gupta et~al.(2016)Gupta, Vedaldi, and Zisserman]{gupta2016synthetic}
Ankush Gupta, Andrea Vedaldi, and Andrew Zisserman.
\newblock Synthetic data for text localisation in natural images.
\newblock In \emph{Proceedings of the IEEE Conference on Computer Vision and
  Pattern Recognition}, pages 2315--2324, 2016.

\bibitem[He et~al.(2016)He, Zhang, Ren, and Sun]{he2016deep}
Kaiming He, Xiangyu Zhang, Shaoqing Ren, and Jian Sun.
\newblock Deep residual learning for image recognition.
\newblock In \emph{Proceedings of the IEEE Conference on Computer Vision and
  Pattern Recognition}, pages 770--778, 2016.

\bibitem[He et~al.(2017)He, Gkioxari, Doll{\'a}r, and Girshick]{he2017mask}
Kaiming He, Georgia Gkioxari, Piotr Doll{\'a}r, and Ross Girshick.
\newblock Mask {R-CNN}.
\newblock In \emph{Proceedings of the IEEE International Conference on Computer
  Vision}, pages 2961--2969, 2017.

\bibitem[Li et~al.(2018)Li, Xu, Liu, Li, and Wang]{li2018jersey}
Gen Li, Shikun Xu, Xiang Liu, Lei Li, and Changhu Wang.
\newblock Jersey number recognition with semi-supervised spatial transformer
  network.
\newblock In \emph{Proceedings of the IEEE Conference on Computer Vision and
  Pattern Recognition Workshops}, pages 1783--1790, 2018.

\bibitem[Lin et~al.(2014)Lin, Maire, Belongie, Hays, Perona, Ramanan,
  Doll{\'a}r, and Zitnick]{lin2014microsoft}
Tsung-Yi Lin, Michael Maire, Serge Belongie, James Hays, Pietro Perona, Deva
  Ramanan, Piotr Doll{\'a}r, and C~Lawrence Zitnick.
\newblock Microsoft {COCO}: Common objects in context.
\newblock In \emph{European Conference on Computer Vision}, pages 740--755.
  Springer, 2014.

\bibitem[Liu and Bhanu(2019)]{liu2019pose}
Hengyue Liu and Bir Bhanu.
\newblock Pose-guided {R-CNN} for jersey number recognition in sports.
\newblock In \emph{Proceedings of the IEEE/CVF Conference on Computer Vision
  and Pattern Recognition Workshops}, pages 0--0, 2019.

\bibitem[Liu and Bhanu(2022)]{liu2022jede}
Hengyue Liu and Bir Bhanu.
\newblock Jede: Universal jersey number detector for sports.
\newblock \emph{IEEE Transactions on Circuits and Systems for Video
  Technology}, 32\penalty0 (11):\penalty0 7894--7909, 2022.

\bibitem[Nady and Hemayed(2021)]{nady2021player}
Ahmed Nady and Elsayed~E Hemayed.
\newblock Player identification in different sports.
\newblock In \emph{VISIGRAPP (5: VISAPP)}, pages 653--660, 2021.

\bibitem[Ren et~al.(2015)Ren, He, Girshick, and Sun]{ren2015faster}
Shaoqing Ren, Kaiming He, Ross Girshick, and Jian Sun.
\newblock Faster {R-CNN}: Towards real-time object detection with region
  proposal networks.
\newblock In \emph{Advances in Neural Information Processing Systems}, pages
  91--99, 2015.

\bibitem[Russakovsky et~al.(2015)Russakovsky, Deng, Su, Krause, Satheesh, Ma,
  Huang, Karpathy, Khosla, Bernstein, Berg, and Fei-Fei]{ILSVRC15}
Olga Russakovsky, Jia Deng, Hao Su, Jonathan Krause, Sanjeev Satheesh, Sean Ma,
  Zhiheng Huang, Andrej Karpathy, Aditya Khosla, Michael Bernstein,
  Alexander~C. Berg, and Li Fei-Fei.
\newblock {ImageNet Large Scale Visual Recognition Challenge}.
\newblock \emph{International Journal of Computer Vision (IJCV)}, 115\penalty0
  (3):\penalty0 211--252, 2015.

\bibitem[Singh et~al.(2021)Singh, Pang, Toh, Huang, Galuba, and
  Hassner]{singh2021textocr}
Amanpreet Singh, Guan Pang, Mandy Toh, Jing Huang, Wojciech Galuba, and Tal
  Hassner.
\newblock {TextOCR}: Towards large-scale end-to-end reasoning for
  arbitrary-shaped scene text.
\newblock In \emph{Proceedings of the IEEE/CVF Conference on Computer Vision
  and Pattern Recognition}, pages 8802--8812, 2021.

\bibitem[Vats et~al.(2021)Vats, Fani, Clausi, and Zelek]{vats2021multi}
Kanav Vats, Mehrnaz Fani, David~A Clausi, and John Zelek.
\newblock Multi-task learning for jersey number recognition in ice hockey.
\newblock In \emph{Proceedings of the 4th International Workshop on Multimedia
  Content Analysis in Sports}, pages 11--15, 2021.

\bibitem[Vats et~al.(2022)Vats, McNally, Walters, Clausi, and
  Zelek]{vats2022ice}
Kanav Vats, William McNally, Pascale Walters, David~A Clausi, and John~S Zelek.
\newblock Ice hockey player identification via transformers and weakly
  supervised learning.
\newblock In \emph{Proceedings of the IEEE/CVF Conference on Computer Vision
  and Pattern Recognition}, pages 3451--3460, 2022.

\bibitem[Vats et~al.(2023)Vats, Walters, Fani, Clausi, and
  Zelek]{vats2023player}
Kanav Vats, Pascale Walters, Mehrnaz Fani, David~A Clausi, and John~S Zelek.
\newblock Player tracking and identification in ice hockey.
\newblock \emph{Expert Systems with Applications}, 213:\penalty0 119250, 2023.

\bibitem[Veit et~al.(2016)Veit, Matera, Neumann, Matas, and
  Belongie]{veit2016coco}
Andreas Veit, Tomas Matera, Lukas Neumann, Jiri Matas, and Serge Belongie.
\newblock {COCO-Text}: Dataset and benchmark for text detection and recognition
  in natural images.
\newblock \emph{arXiv preprint arXiv:1601.07140}, 2016.

\bibitem[Wieczorek et~al.(2021)Wieczorek, Rychalska, and
  Dabrowski]{wieczorek2021unreasonable}
Mikolaj Wieczorek, Barbara Rychalska, and Jacek Dabrowski.
\newblock On the unreasonable effectiveness of centroids in image retrieval.
\newblock In \emph{Neural Information Processing: 28th International
  Conference, ICONIP 2021, Sanur, Bali, Indonesia, December 8--12, 2021,
  Proceedings, Part IV 28}, pages 212--223. Springer, 2021.

\bibitem[Xu et~al.(2022)Xu, Zhang, Zhang, and Tao]{xu2022vitpose}
Yufei Xu, Jing Zhang, Qiming Zhang, and Dacheng Tao.
\newblock {ViTPose}: Simple vision transformer baselines for human pose
  estimation.
\newblock \emph{Advances in Neural Information Processing Systems},
  35:\penalty0 38571--38584, 2022.

\bibitem[Zhao et~al.(2020)Zhao, Li, and Chen]{zhao20}
Yingnan Zhao, Zihui Li, and Kua Chen.
\newblock A method for tracking hockey players by exploiting multiple
  detections and omni-scale appearance features.
\newblock \emph{Project Report}, 2020.

\bibitem[Zheng et~al.(2015)Zheng, Shen, Tian, Wang, Wang, and
  Tian]{zheng2015scalable}
Liang Zheng, Liyue Shen, Lu Tian, Shengjin Wang, Jingdong Wang, and Qi Tian.
\newblock Scalable person re-identification: A benchmark.
\newblock In \emph{Proceedings of the IEEE international conference on computer
  vision}, pages 1116--1124, 2015.

\end{thebibliography}
}


\end{document}